\documentclass{article}

\usepackage[final]{nips_2017}

\usepackage[utf8]{inputenc} 
\usepackage[T1]{fontenc}    
\usepackage{hyperref}       
\usepackage{url}            
\usepackage{booktabs}       
\usepackage{amsfonts}       
\usepackage{nicefrac}       
\usepackage{microtype}      
\usepackage{amsmath}
\usepackage[pdftex]{graphicx}
\usepackage{rotating}
\usepackage{multirow}

\title{SUSAN: Segment Unannotated image Structure using Adversarial Network}

\author{Fang Liu \\
Department of Radiology\\
University of Wisconsin-Madison\\
Madison, Wisconsin USA, 53705-2275 \\
  \texttt{leoliuf@gmail.com} \\
}

\begin{document}

\maketitle

\begin{abstract}
Purpose: To describe and evaluate a segmentation method using joint adversarial and segmentation convolutional neural network (CNN) to achieve accurate segmentation using unannotated magnetic resonance (MR) image datasets.
Methods: A segmentation pipeline was built using joint adversarial and segmentation network. A CNN technique called cycle-consistent generative adversarial network (CycleGAN) was applied as the core of the method to perform unpaired image-to-image translation between different MR image datasets. A joint segmentation network was incorporated into the adversarial network to obtain additional functionality for semantic segmentation. The fully-automated segmentation method termed as SUSAN was tested for segmenting bone and cartilage on two clinical knee MR image datasets using images and annotated segmentation masks from an online publicly available knee MR image dataset. The segmentation results were compared using quantitative segmentation metrics with the results from a supervised U-Net segmentation method and two registration methods. The Wilcoxon signed-rank test was used to evaluate the value difference of quantitative metrics between different methods.
Results: The proposed method SUSAN provided high segmentation accuracy with results comparable to the supervised U-Net segmentation method (most quantitative metrics having \textit{p}>0.05) and significantly better than a multi-atlas registration method (all quantitative metrics having \textit{p}<0.001) and a direct registration method (all quantitative metrics having \textit{p}<0.0001) for the clinical knee image datasets. SUSAN also demonstrated the applicability for segmenting knee MR images with different tissue contrasts.
Conclusion: SUSAN performed rapid and accurate tissue segmentation for multiple MR image datasets without the need for sequence specific segmentation annotation. The joint adversarial and segmentation network and training strategy have promising potential applications in medical image segmentation.

\end{abstract}

\section{INTRODUCTION}
Segmentation of magnetic resonance (MR) images is a fundamental step in many medical imaging based applications. Traditionally, image segmentation is performed by having experienced users scroll through stacks of two-dimensional (2D) images and manually segmenting regions-of-interest (ROIs) among adjacent tissues.  However, manual segmentation is time-consuming and is influenced by the level of human expertise and errors due to distraction and fatigue associated with human interpretation(1–4). Therefore, manual segmentation is subject to inter- and intra- observer variability which likely leads to inconsistent segmentation results(3,4).  There has been much recent interest in developing semi- and fully-automated techniques for segmenting MR images(5). The majority of recently proposed methods for fully-automated segmentation have utilized model-based and atlas-based approaches(5,6). Although these methods have shown promising results, both approaches rely on a priori knowledge of object shapes and thus might perform poorly in situations in which there is high subject variability and significant differences of local features. In addition, these methods require high computation cost which results in relatively long segmentation times.

Recent implementation of deep convolutional neural networks (CNNs) in image processing has been shown to have significant impacts on medical image segmentation (7). Deep CNN-based methods have achieved state-of-the-art performance in many medical image segmentation tasks including segmenting brain tumors(8,9), tissues (10,11), and multiple sclerosis lesions (12), cardiac (13,14), liver(15), and lung(16) tissues, and musculoskeletal tissues such as bone and cartilage(17–19). On the other hand, medical image segmentation is typically seen as a multi-class labeling problem which is closely related to the supervised semantic segmentation described in most segmentation CNN studies. In particular, convolutional encoder-decoder (CED) networks have proven to be highly efficient in the medical image domain. This type of network typically consists of a paired encoder and decoder where the encoder performs image compression and feature extraction and the decoder reconstructs pixel-wise classification labels using encoder outputs. Ronneberger et al. (20) developed U-Net which has a 2D CED structure with skip connections between the encoder and decoder. The U-Net transfers feature maps from the encoder to the decoder and concentrates them to obtain up-sampled feature maps through deconvolution. U-Net was first proposed for segmenting neuronal structures in electron microscopy and was later adapted for many medical image segmentation tasks. Badrinarayanan et al. (21) proposed a 2D CED called SegNet which is built upon the deep structure of the VGG16 network (22) and features a unique up-sampling approach in the decoder using pre-stored max-pooling indices from the encoder. This network offers an efficient alternative to deconvolution for recovering high resolution image features and achieved top performance in multiple segmentation challenges. Expanding on the capabilities of 2D CEDs, such as U-Net and SegNet, recently proposed three-dimensional (3D) CEDs extend convolutional kernels into the slice dimension for volumetric image data and attempt to incorporate full spatial information for improved segmentation performance(23,24). Other segmentation networks using multi-scale and multi-patch based structures have also been proposed and have proven to be quite useful for segmenting 3D image datasets(9,25,26). More recently, further improvement for CNN-based image segmentation was achieved by using adversarial training, where a dedicated CNN network was introduced to correct generated segmentation maps from the ground truth(27). A few pilot studies demonstrated great performance using adversarial training for segmenting brain lesions(28), structures(29), and prostate cancer(30) on MR images, chest organs(31) on X-ray images, and breast cancer on histopathology images(32).

Network training of segmentation CNNs typically requires images and paired annotation data representing pixel-wise tissue labels referred to as masks. The pixel-wise correlation between image pixels and tissue masks is used for supervised training of the segmentation CNNs for learning useful image features. However, the supervised training of highly efficient CNNs with deeper structure and more network parameters requires a large amount of training images and paired tissue masks. Moreover, the creation of tissue masks typically requires individuals with medical expertise to annotate a large number of training image datasets which could be extremely expensive and time consuming(33). Although a trained segmentation CNN may perform well for one type of MR sequence, the applicability of the CNN for segmenting the same tissues on images acquired using other MR sequences is typically poor. Therefore, it is necessary to retrain the CNN using new annotation data specific to each MR sequence. Thus, there is great need to develop a generalized CNN-based segmentation method which would be applicable for a wide variety of MR image datasets with different tissue contrasts.

The purpose of our study was to develop and evaluate a generalized CNN-based method for fully-automated segmentation of different MR image datasets using a single set of annotated training data. A technique called cycle-consistent generative adversarial network (CycleGAN) (34) is applied as the core of the proposed method to perform image-to-image translation between MR image datasets with different tissue contrasts. A segmentation network is incorporated into the adversarial network to obtain additional segmentation functionality. We termed the proposed method as SUSAN standing for Segmenting Unannotated image Structure using Adversarial Network and evaluated SUSAN for segmenting bone and cartilage on two clinical knee MR image datasets acquired at our institution using only a single set of annotated data from a publicly available knee MR image dataset.

\section{THEORY}
\subsection{Adversarial Network for Image-to-Image Translation}

Our work is in line with the method of CycleGAN which was recently proposed for unpaired image-to-image translation for natural images (34). In CycleGAN, images from two image domains can be translated to exchange image contrasts, features and patterns (e.g. translate horse into zebra, apple into orange and vice versa) using two key techniques including Cycle Consistency and Generative Adversarial Newark (GAN). Our work is closely related to the basic framework of CycleGAN. The main concept of SUSAN is to translate the reference images which have high quality segmentation annotation into the target images which have no segmentation annotation. Our hypothesis is that given successful translation from the reference image contrast to the target image contrast, the annotation data used to train the reference images can be applied to train the target images using supervised learning.

\begin{figure}[h]
  \centering
  \includegraphics[width=1.0\linewidth]{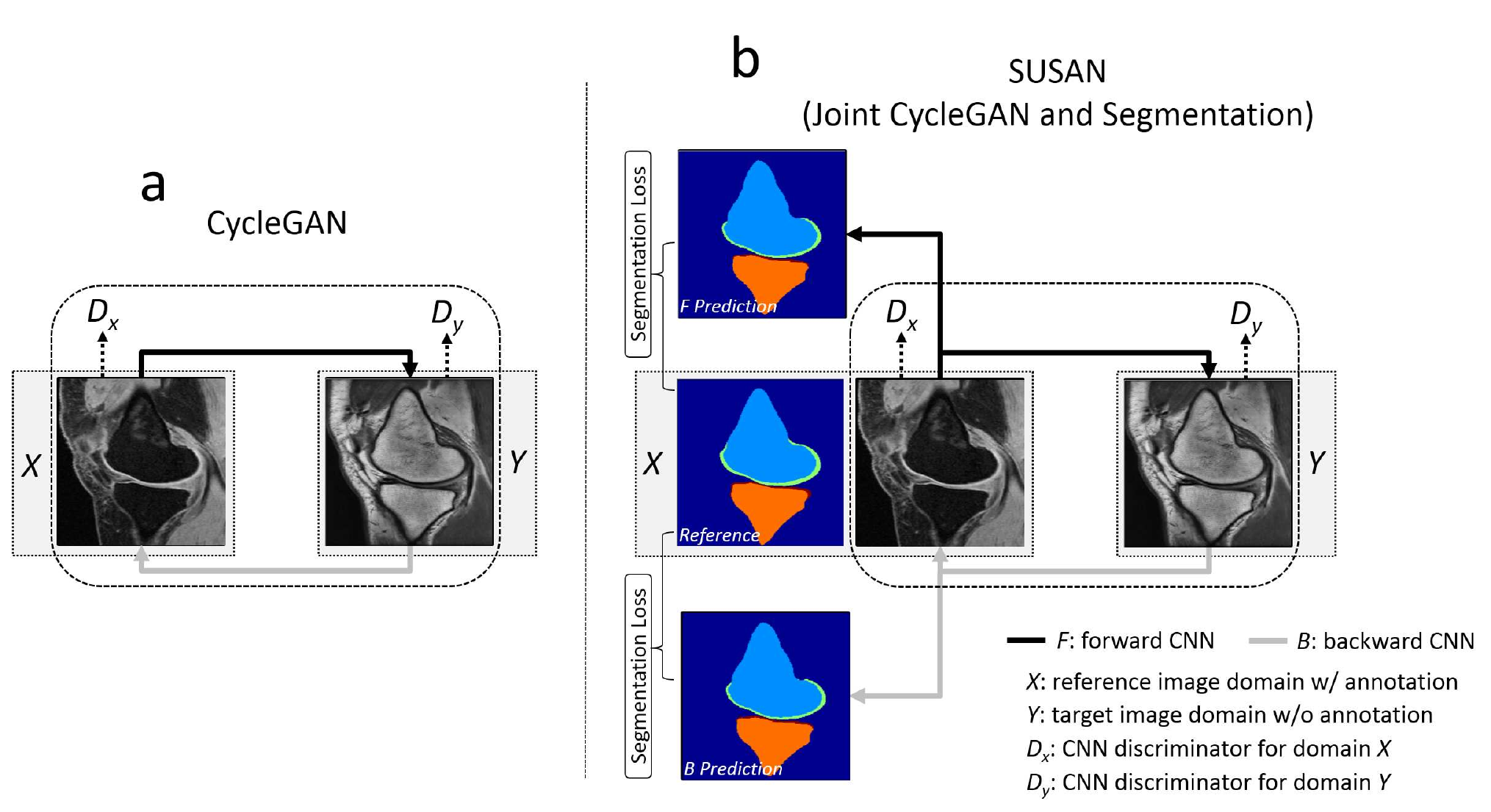}
  \caption{\textbf{a)} The CycleGAN model contains two CNNs, \textit{F} and \textit{B}, for forward and backward mapping between image domain \textit{X} (with annotation) and \textit{Y} (without annotation), and associated adversarial discriminators $D_X$ and $D_Y$.  $D_Y$ encourage \textit{F} to translate images from domain \textit{X} into outputs indistinguishable from domain \textit{Y}, and vice versa for $D_X$ and \textit{B}. The cycle consistency loss is used to enforce the idea that if an image is translated from one domain to the other and back again, the image should look identical to its original version. \textbf{b)} SUSAN: the proposed joint CycleGAN and segmentation model incorporates an additional segmentation branch from the mapping CNNs. Joint training is performed for image translation between \textit{X} and \textit{Y}, and image segmentation for all image \textit{x} from domain \textit{X} and synthetic $F(x)$ which have origins from domain \textit{X} using segmentation loss.}
 \label{fig1}
\end{figure}

Transitivity is a useful tool for many computer science applications including human language translation(35,36), dense semantic alignment(37), image co-segmentation(38) and 3D shape matching(39). In the CycleGAN setup (Figure \ref{fig1}a), the idea of cycle consistency is to apply transitivity to force forward-backward data cycle consistency between two image domains. Mathematically, the minimization of cycle consistency loss is to learn forward (\textit{F}) and backward (\textit{B}) CNN mapping functions:

\begin{equation} \label{eq1}
\begin{array}{l}
F:X \to Y\\
B:Y \to X
\end{array}
\end{equation}

between two image domains \textit{X} and \textit{Y}, so that if an image is translated from one domain to the other and back again, the image should look identical to the original image. Therefore, for forward cycle consistency, there should be ${x \to F(x) \to B(F(x)) \approx x}$ and for backward cycle consistency, there should be ${y \to B(y) \to F(B(y)) \approx y}$. The cycle consistency loss term can be formulated as 

\begin{equation} \label{eq2}
{L_{cyc}}(F,B) = {\mathbb{E}_{x \to P(x)}}\left[ {{{\left\| {B(F(x)) - x} \right\|}_N}} \right] + {\mathbb{E}_{y \to P(y)}}\left[ {{{\left\| {F(B(y)) - y} \right\|}_N}} \right]
\end{equation}

where ${{\mathbb{E}_{i \to P(i)}}\left[  \cdot  \right]}$ is the expectation of a function and the data distribution of image \textit{i} in domain \textit{I} is denoted as ${i \to P\left( i \right)}$. The ${{\left\|  \cdot  \right\|_N}}$ is typically chosen to be ${l_1}$ norm for image-to-image translation (34).

GAN is another key concept for CycleGAN. Many recent GAN studies have achieved impressive results in a variety of image subspecialties including image inpainting (40), text to image synthesis(41), and image generation(42,43). The idea of GAN is to use an \textit{adversarial loss} to force CNN mapping function generating synthetic images that are indistinguishable from the real images. In the current study, adversarial loss can be incorporated into the CNN training as a joint loss term. For example, for the CNN mapping function ${F:X \to Y}$, a multiple layer CNN discriminator ${{D_Y}(y)}$ is defined to identify real versus synthetic images. Mathematically, this discriminator outputs a scalar representing the probability that y comes from the real domain \textit{Y} rather than the forward mapping output . The adversarial loss term is thus formatted as

\begin{equation} \label{eq3}
L_{gan}^f(F,{D_Y}) = {\mathbb{E}_{y \to P(y)}}\left[ {\log {D_Y}(y)} \right] + {\mathbb{E}_{x \to P(x)}}\left[ {\log (1 - {D_Y}(F(x)))} \right]
\end{equation}

In addition, the CycleGAN also introduced a CNN discriminator ${{D_X}(x)}$ for the backward mapping with adversarial loss as:

\begin{equation} \label{eq4}
L_{gan}^b(B,{D_X}) = {\mathbb{E}_{x \to P(x)}}\left[ {\log {D_X}(x)} \right] + {\mathbb{E}_{y \to P(y)}}\left[ {\log (1 - {D_X}(B(y)))} \right]
\end{equation}

The full objective function for CycleGAN is given as:

\begin{equation} \label{eq5}
L(F,B,{D_X},{D_Y}) = {\lambda _{cyc}}{L_{cyc}}(F,B) + {\lambda _{gan}}\left( {L_{gan}^f(F,{D_Y}) + L_{gan}^b(B,{D_X})} \right)
\end{equation}

where ${{\lambda _{cyc}}}$ and ${{\lambda _{gan}}}$ are weight factors for the cycle consistency loss and adversarial loss term, respectively, to balance the data fidelity of image translation and the GAN quality. The full objective function is trained in a two-player minimax game, namely \textit{F} and \textit{B} aim to minimize this loss function against the adversary $D_X$ and $D_Y$ that try to maximize it as

\begin{equation} \label{eq6}
\tilde F,\tilde B = \arg \mathop {\min }\limits_{F,B} \mathop {\max }\limits_{{D_X},{D_Y}} L(F,B,{D_X},{D_Y})
\end{equation}

In other words, \textit{F} and \textit{B} try to generate synthetic images that look similar to real images, while discriminator $D_X$ and $D_Y$ try to distinguish synthetic images from real images. In theory, successful training of this network can result in mapping CNNs capable of generating synthetic images indistinguishable from real images in the original domain(34,44). Namely, the translated images in each image domain just look like real images from that domain.

\subsection{Joint Segmentation Network}
To make use of image-to-image translation, SUSAN incorporates additional segmentation networks into the CycleGAN structure for jointly training image translation and segmentation (Figure \ref{fig1}b). Our hypothesis is that the image translation can be augmented by adding supervised segmentation information and that jointly training image translation and segmentation can produce improved results. To formulate the problem, the CNN mapping function is modified to create a dual-output network. Namely, the mapping function is designed to not only output translated images but also full size tissue segmentation masks. Therefore, Eq.\ref{eq1} can be rewritten as

\begin{equation} \label{eq7}
\begin{array}{l}
F:X \to (Y,{M_f})\\
B:Y \to (X,{M_b})
\end{array}
\end{equation}

where $M_f$ and $M_b$ are output segmentation masks for the forward and backward CNN, respectively. It is assumed that all the images in image domain \textit{X} have high quality segmentation masks $M_x$ which can be used for supervised training. The segmentation loss is incorporated into the full objective function as

\begin{equation} \label{eq8}
{L_{seg}}(F,B) = {\mathbb{E}_{x \to P(x)}}\left[ {\ell (F(x),{M_x})} \right] + {\mathbb{E}_{x \to P(x)}}\left[ {\ell (B(F(x)),{M_x})} \right]
\end{equation}

where ${\ell ( \cdot )}$ is a loss metric for evaluating pixel-wise similarity between the output mask and the ground truth mask, e.g. multi-class cross entropy (45). According to Eq.\ref{eq8}, supervised segmentation training is performed for all image \textit{x} in domain \textit{X} and all translated synthetic image $F(x)$  which have origins from image domain \textit{X}. The full objective function incorporating segmentation loss can be extended from Eq.\ref{eq5} into

\begin{equation} \label{eq9}
L(F,B,{D_X},{D_Y}) = {\lambda _{cyc}}{L_{cyc}}(F,B) + {\lambda _{seg}}{L_{seg}}(F,B) + {\lambda _{gan}}\left( {L_{gan}^f(F,{D_Y}) + L_{gan}^b(B,{D_X})} \right)
\end{equation}

where ${{\lambda _{seg}}}$ is a weight factor for the segmentation term. Note that once the joint training process is complete, for image domain \textit{X}, the final segmentation CNN is simply given by the forward CNN as

\begin{equation} \label{eq10}
{M_f} = F\left( x \right),x \in X
\end{equation}

Likewise, for image domain \textit{Y} which has no explicit annotated segmentation mask, the final segmentation result is given by the backward CNN output as 

\begin{equation} \label{eq11}
{M_b} = B\left( y \right),y \in Y
\end{equation}

\section{METHODS}
\subsection{Network Implementation}
\begin{figure}[h]
  \centering
  \includegraphics[width=0.85\linewidth]{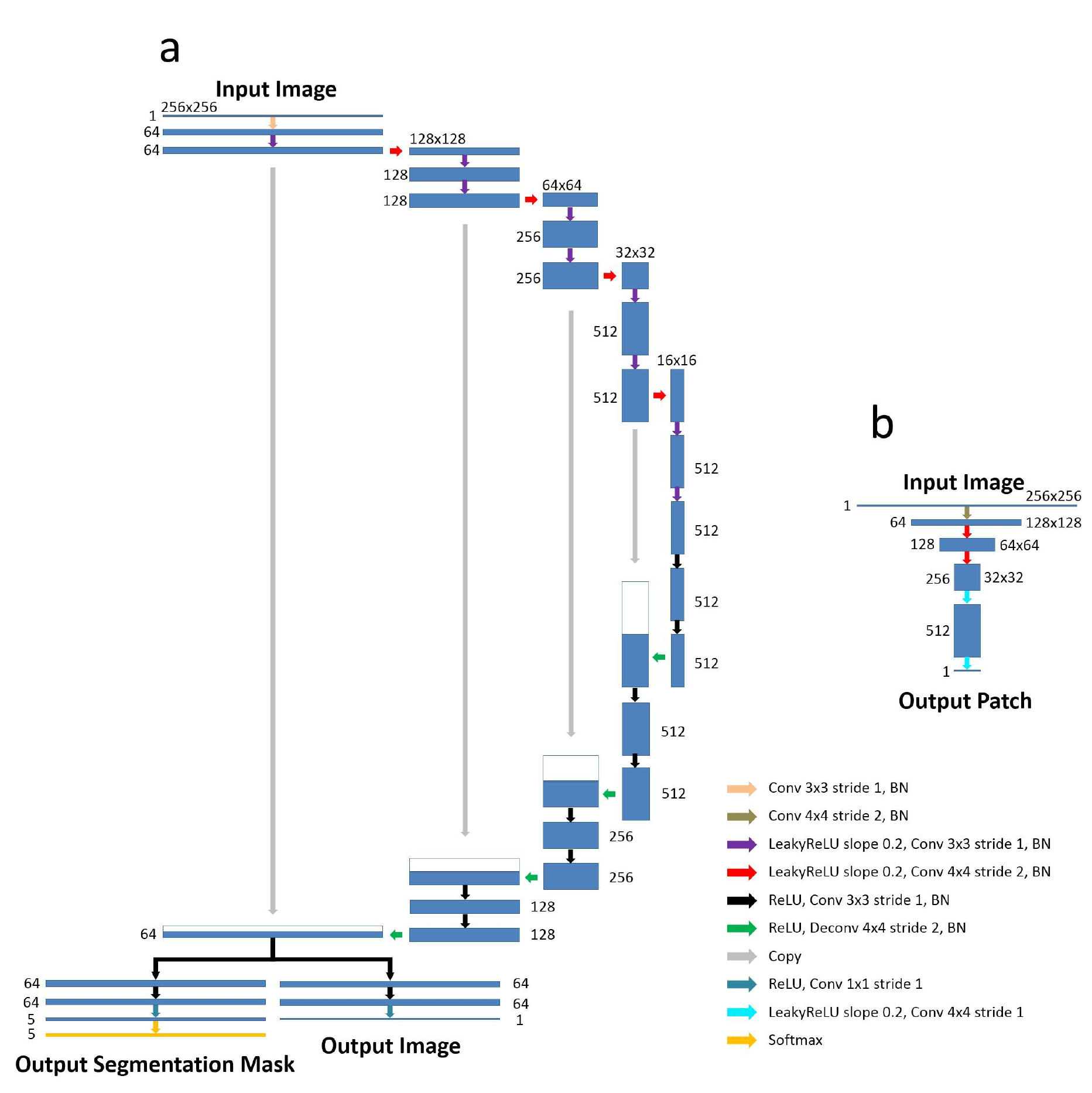}
  \caption{A schematic illustration of the CNN architectures used in current SUSAN. \textbf{a)} The R-Net modified from standard U-Net structure is used for the CNN mapping function \textit{F} and \textit{B}. The R-Net allows two outputs each of which performs image segmentation and image translation, separately. The joint portion of the R-Net for segmentation and translation branch enables sharing image features during network training. \textbf{b)} The discriminator CNN designed in PatchGAN is used for $D_X$ and $D_Y$. This network outputs image patches with reduced image size which will be used for differentiating real versus synthetic images in the adversarial training process. (Abbreviation: Conv: Convolution Layer; BN: Batch Normalization; ReLU: Rectified Linear Unit; LeakyReLU: Leaky Rectified Linear Unit; Deconv: Transpose Convolution Layer; Softmax: Softmax Layer)}
 \label{fig2}
\end{figure}

A U-Net architecture (20) was adapted from a GAN-based image-to-image translation study (46) for performing the CNN mapping functions (i.e. \textit{F} and \textit{B}) between the two image domains. This U-Net structure is composed of an encoder network and a decoder network. The encoder is used to achieve efficient data compression while probing robust and spatial invariant image features. A decoder network with a mirrored structure of the encoder is applied following the encoder network output for restoring desirable image features. Multiple symmetric shortcut connections are added to transfer features from the encoder to the decoder to enhance mapping performance. Such a CNN structure has shown impressive results for image translation and segmentation in many recent studies(20,46,47). In the current study, the U-Net structure is modified to enable dual outputs, and this new design is referred to as R-Net. Namely, the network is bifurcated following the last up-sampling layer in the decoder. One segmentation branch of R-Net uses a multi-class soft-max classification layer as the final layer (48) which produces class probabilities for each individual pixel at the same image resolution as the input image. The image translation branch of R-Net uses a convolution layer as the last layer to generate gray scale image also matching the input image resolution. An illustration of the R-Net is shown in Figure \ref{fig2}a. 

Similar to the original CycleGAN work, the network architecture developed in PatchGAN (46) is used for discriminator networks (i.e. $D_X$ and $D_Y$), which aim to classify whether overlapping image patches are real or fake in the adversarial process. An illustration of PatchGAN is shown in Figure \ref{fig2}b. Such a patch-level discriminator architecture has the advantage of fewer parameters allowing for more efficient network training and has been shown in many recent GAN studies to have impressive results for differentiating real versus synthetic images(46,49,50).

\subsection{Image Datasets}
Three knee MR image datasets were used to evaluate SUSAN. The reference image dataset consisted of the online knee MR images available from The Segmentation of Knee Images 2010 (SKI10, www.ski10.org) workshop competition hosted by the Medical Image Computing and Computer Assisted Intervention (MICCAI) conference in 2010 (51). The image dataset in SKI10 consisted of 60 sagittal fat-suppressed T1-weighted spoiled gradient-echo (T1-SPGR) knee images, all of which had high quality multi-class tissue masks manually generated by segmentation experts with the following values: 0=background, 1=femur, 2=femoral cartilage, 3=tibia, and 4=tibial cartilage.

\begin{table}[t]
  \caption{Imaging parameters for the sagittal fat-suppressed T1-weighted spoiled gradient-echo (T1-SPGR) sequence in the SKI10 image dataset and the two clinical MR sequences including sagittal fat-suppressed T2-weighted fast spin-echo (T2-FSE) and sagittal proton density-weighted fast spin-echo (PD-FSE).}
   \label{tb1}
  \centering
  \begin{tabular}{llll}
    \toprule
    Sequence  &  T2-FSE     &  PD-FSE   &  T1-SPGR  \\
    \midrule
    Field Strength [T] 	 & 3.0 & 3.0   &  1, 1.5, 3.0  \\
    TR [ms]              & 4680  & 1900   &  Unavailable  \\
    TE [ms] 		     & 80 & 19   &  Unavailable \\
    Flip Angle [degree]   & 90 & 90   &  Unavailable  \\
    Pixel Bandwidth [Hz]   & 163 & 122  &  Unavailable  \\
    Echo Train Length   & 20 & 4   &  N/A  \\
    Field of View [cm]   & 14  & 14   &  14  \\
    Slice Thickness [mm]   & 3  & 2  &  1\\
    Slice Gap [mm]   & 0.5  & 0.5   &  N/A  \\
    Number of Slices   & 23$\sim$30  & 32$\sim$42  &  90$\sim$120  \\
    Matrix Size   & 512$\times$512  & 512$\times$512  &  360$\times$300 \\
    \bottomrule 
  \end{tabular}
\end{table}

Two clinical knee MR image datasets acquired at our institution were used as target images. The study was performed in compliance with Health Insurance Portability and Accountability Act (HIPAA) regulations, with approval from our Institutional Review Board, and with a waiver of written informed consent. Sagittal knee image datasets acquired with two MR sequences were retrospectively obtained on patients undergoing a clinical knee MR examination at our institution using a 3T scanner (Discovery MR750, GE Healthcare) and 8-channel phased-array extremity coil.  The first image dataset consisted of a sagittal fat-suppressed T2-weighted fast spin-echo (T2-FSE) sequence acquired on 60 patients. The second image dataset consisted of a sagittal proton density-weighted fast spin-echo (PD-FSE) sequence acquired on 60 patients. The imaging parameters for both MR sequences are summarized in Table \ref{tb1}. Manual bone and cartilage segmentation of the clinical knee images was performed by a musculoskeletal research scientist with 8 years of segmentation experience. Note that the manual segmentation for the clinical image datasets was used only for ground truth comparison and was not included in the network training for SUSAN. Multi-class tissue masks were created for the clinical knee image datasets using the same labelling values as used for the SKI10 image dataset.

\subsection{Network Training}

All the input 2D images from the clinical knee MR image datasets were first cropped to enclose as much of the knee joint as possible while removing excessive image background, leading to an approximate 512$\times$400 image matrix size. Due to the GPU global memory limit, all the images were further resampled to 256$\times$256 matrix size using bilinear interpolation before they were sent to the network for training and evaluation. Image normalization was also performed for each subject by subtracting the mean value of the entire image and then normalizing by the standard deviation of the image signal intensity. When training the network, the network weights were initialized using the initialization scheme of He et al. (52) and updated using Adam algorithm (53) with a fixed learning rate of 0.0002 and trained in a mini-batch manner with three image slices in a single mini-batch. Multi-class cross entropy loss was applied for the segmentation branch (in Eq.\ref{eq8}) and the ${l_1}$ loss for the image translation branch (in Eq.\ref{eq2}) in the R-Net. During training iteration, a two-step training strategy was applied where CNN mapping functions (i.e. \textit{F} and \textit{B}) and adversarial discriminators (i.e. $D_X$ and $D_Y$) were updated separately in an alternating manner. A default set of parameters for the weight factors in the full objective function was empirically selected and included
${\lambda _{cyc}}$= 10, ${\lambda _{gan}}$=1 and ${\lambda _{seg}}$= 5 in the Eq.\ref{eq9} for the clinical knee image datasets. To investigate the influence of different weights for the segmentation branch on the performance of image translation and subsequent segmentation, the experiments were also performed for ${\lambda _{seg}}$= 0.5 and 10, respectively, while maintaining other parameters the same.

For the network training of SUSAN, a split of 50/10 subjects (5506/1083 slices) in the SKI10 image dataset and a split of 35/5 randomly selected subjects in the PD-FSE (1343/201 slices) and T2-FSE (937/146 slices) image datasets were used for training and validation, respectively, while the remaining 20 subjects in the PD-FSE and T2-FSE image datasets were used for hold-out evaluation. A total iteration steps corresponding to 20 epochs for the SKI10 image dataset were carried out for training convergence. The best model was selected in which the calculated loss was the lowest for the set of validation images. To compare SUSAN with a standard supervised method, direct U-Net training for tissue segmentation was performed using the clinical knee image datasets, and the corresponding multi-class tissue masks created using manual segmentation. The supervised learning was performed by removing the image translation branch from the R-Net while keeping the aforementioned training procedure the same. To compare SUSAN with state-of-the-art conventional segmentation methods, a multi-atlas registration algorithm from the Knee Segmentation and Registration Toolkit (KSRT, https://bitbucket.org/marcniethammer/ksrt)(54) was evaluated. Multiple atlases were built from all 60 subjects in SKI10 image dataset and the registration workflow and parameters were kept the same as the default setting implemented in the source code and stated in the original paper(54). In addition, to compare SUSAN with direct registration approach between different image contrasts, a registration algorithm using Elastix software and registration parameters (http://elastix.isi.uu.nl/wiki.php) as described in (55) was also evaluated. The segmentation of bone and cartilage was obtained for the clinical datasets by directly registering the images into each individual SKI10 subject image as one template. The segmentation accuracy of SUSAN, the supervised U-Net method, the multi-atlas registration and the direct registration method (value reported using the best segmentation from all templates in SKI10 datasets) was evaluated on the 20 hold-out test subjects in the PD-FSE and T2-FSE image datasets.

SUSAN was implemented in Python language (v2.7, Python Software Foundation, Wilmington, Del). The CNNs were designed using the Keras package (56) running Tensorflow computing backend (57) at a 64-bit Ubuntu Linux system. All training and evaluation were performed on a computer server with an Intel Xeon W3520 quad-core CPU, 32 GB DDR3 RAM, and one Nvidia GeForce GTX 1080Ti graphic card with total 3584 CUDA cores, and 11GB GDDR5 RAM.

\subsection{Evaluation of Synthetic Images}

To evaluate the quality of synthetic images, the FCN-score, a quantitative metric used in the original CycleGAN paper (34) was implemented for image assessment. While direct evaluation of the generative model is challenging, the idea is to use a pre-trained semantic segmentation classifier to measure the discriminability of synthetic images against the ground truth images. The assumption is that the segmentation classifier trained on the ground truth images should be able to segment synthetic images at a high accuracy if the synthetic images are realistic(34,46). More specifically, a FCN-8s network (45) was adapted for the segmentation classifier, and was trained on the real PD-FSE and T2-FSE dataset, respectively, using the same aforementioned training procedure. The trained model was then applied to segment 20 synthetic PD-FSE and T2-FSE images and 20 real hold-out clinical knee images for comparison, respectively. The Per-pixel accuracy was reported as FCN-score with the definition as follows

\begin{equation} \label{eq12}
ACC = \frac{{\sum\limits_i {{n_{ii}}} }}{\sum\limits_i {t_i}}
\end{equation}

where ${n_{ii}}$ is the number of pixels of class \textit{i} correctly predicted to belong to class \textit{i}, and $t_i$ is the total number of pixels of class \textit{i}. The FCN-score implementation in this study was adapted from the online code at https://github.com/phillipi/pix2pix/tree/master/scripts/eval\_cityscapes.

\subsection{Evaluation of Segmentation Accuracy}

Quantitative metrics were used to evaluate the accuracy of the segmentation methods on the different clinical knee MR image datasets. To evaluate the volumetric segmentation accuracy, the Dice coefficient (DC) was used for bone and cartilage and was defined as 

\begin{equation} \label{eq13}
\rm{DC} = \frac{{2\left| {S \cap R} \right|}}{{\left| S \right| + \left| R \right|}}
\end{equation}

where \textit{S} and \textit{R} represented the CNN segmentation and the manual segmentation ground truth, respectively. The Dice coefficient ranges between 0 and 1 with a value of 1 indicating a perfect segmentation and a value of 0 indicating no overlap at all. The volumetric overlap error (VOE) was also calculated to evaluate the accuracy of cartilage segmentation. The VOE was defined as 

\begin{equation} \label{eq14}
\rm{VOE} = 1 - \frac{{\left| {S \cap R} \right|}}{{\left| {S \cup R} \right|}}
\end{equation}

with a smaller VOE value indicating a more accurate segmentation. The VOE values were calculated within a ROI drawn in each of three consecutive central slices on the medial and lateral tibial plateau, and medial and lateral femoral condyles. To evaluate the surface overlap between the segmented masks and the ground truth, the Average Symmetric Surface Distance (ASSD) was calculated for bone and cartilage. The ASSD was defined as 

\begin{equation} \label{eq15}
\rm{ASSD} = \frac{{\mathop \sum \nolimits_{s \in \partial \left( S \right)} \mathop {\min }\limits_{r \in \partial \left( R \right)} \left\| {s - r} \right\| + \mathop \sum \nolimits_{r \in \partial \left( R \right)} \mathop {\min }\limits_{s \in \partial \left( S \right)} \left\| {r - s} \right\|{\rm{\;}}}}{\left| {\partial \left( S \right)} \right| + \left| {\partial \left( R \right)} \right|}
\end{equation}

where $\partial ( \cdot )$ means the boundary of the segmentation set. A small ASSD value typically indicates similar surface boundaries and great surface overlap, thereby reflecting a more accurate segmentation. For statistical analysis, a paired non-parametric Wilcoxon signed-rank test was used to compare the DC, VOE and ASSD values between SUSAN, the supervised U-Net method, and two registration methods at a pre-defined significance level of \textit{p}<0.05.

\section{RESULTS}
The overall training time required for SUSAN was approximate 6.7 hours for each clinical knee MR image dataset given the computing hardware in the current study.  However, once the training was complete, fully-automated segmentation was rapid with a mean computing time of 0.2 min for all image slices in the PD-FSE and T2-FSE image datasets. In contrast, the multi-atlas registration method took a mean computing time of 5.2 hours for all image slices in the PD-FSE and T2-FSE image datasets. The direct registration method took less time with a mean computing time of 0.5 hours for all image slices in the PD-FSE and T2-FSE image datasets using one template.

\begin{figure}[h]
  \centering
  \includegraphics[width=0.5\linewidth]{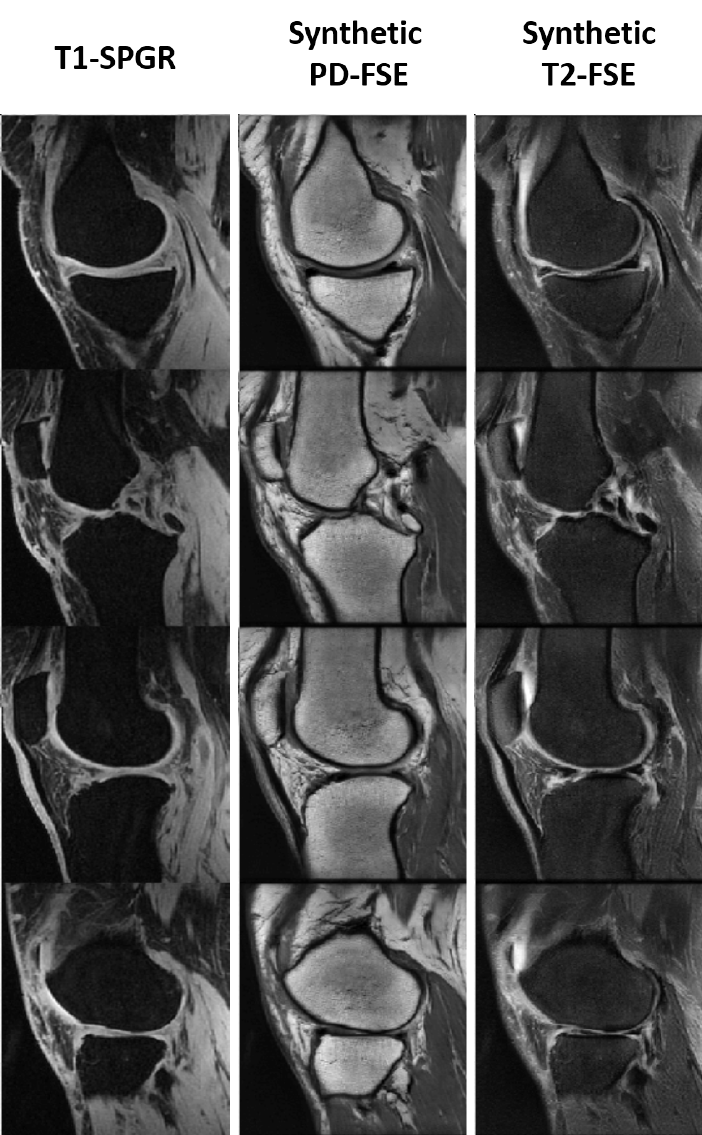}
  \caption{Example of unpaired image-to-image translation for converting SKI10 T1-SPGR image contrast to the PD-FSE and T2-FSE image contrasts (denoted as synthetic PD-FSE and T2-FSE). Note that the subject was randomly selected for demonstration. Despite the dramatic differences in tissue contrasts between the three image datasets, SUSAN was capable of successfully translating the varying MR contrasts.}
 \label{fig3}
\end{figure}

\begin{figure}[h]
  \centering
  \includegraphics[width=0.7\linewidth]{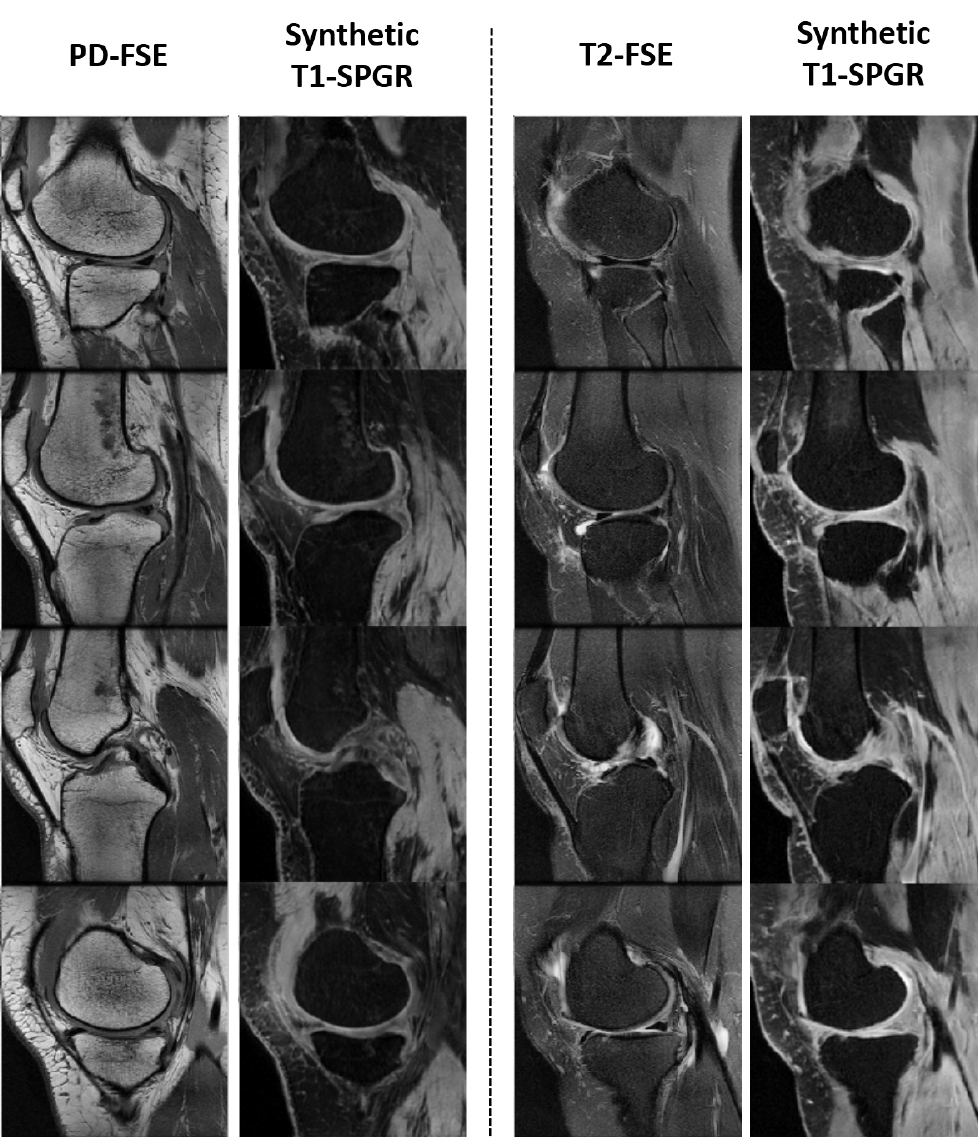}
  \caption{Example of unpaired image-to-image translation for converting the PD-FSE and T2-FSE image contrasts to SKI10 T1-SPGR image contrast (denoted as synthetic T1-SPGR). Note that the subject was randomly selected from the individual image datasets for demonstration. Despite the dramatic differences in tissue contrasts between the three image datasets, SUSAN was capable of successfully translating the varying MR contrasts.}
 \label{fig4}
\end{figure}

\begin{figure}[h]
  \centering
  \includegraphics[width=0.7\linewidth]{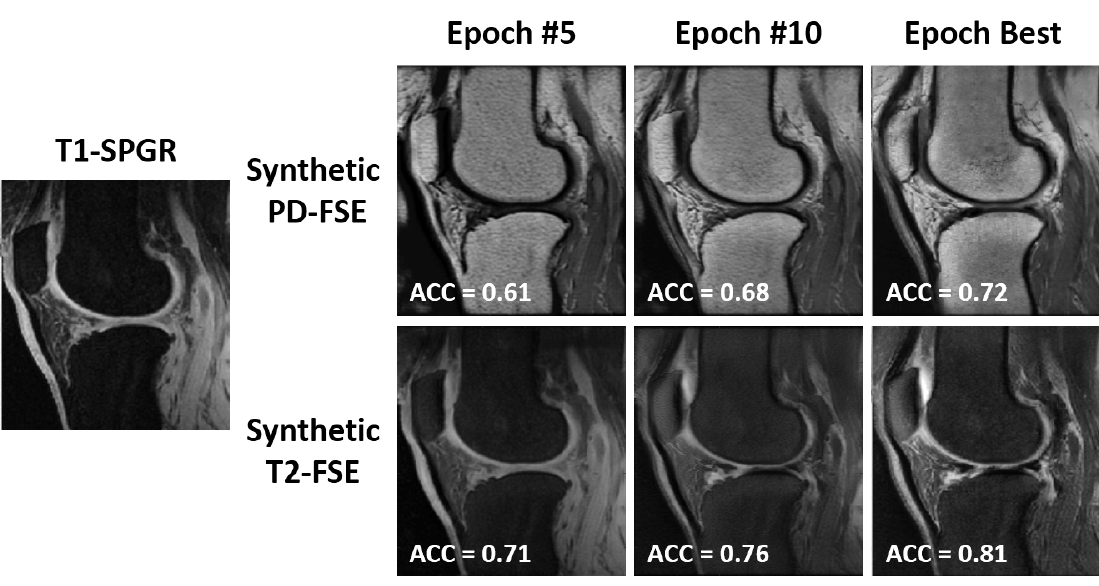}
  \caption{Example of synthetic PD-FSE and T2-FSE images from a SKI10 T1-SPGR image at different epochs. The FCN-scores were calculated using ACC metric to illustrate quantitative assessment of the image quality.}
 \label{fig5}
\end{figure}

Figure \ref{fig3} and \ref{fig4} show an example of unpaired image-to-image translation for converting the SKI10 T1-SPGR image contrast to the PD-FSE and T2-FSE image contrasts (denoted as synthetic PD-FSE and T2-FSE) and vice versa. Note that in Figure \ref{fig3} and \ref{fig4}, the subject was randomly selected from the individual image datasets for demonstration. The T1-SPGR images had low fat signal in bone and subcutaneous soft tissue due to fat suppression and bright cartilage and muscle signal. The PD-FSE images had high fat signal in bone and subcutaneous soft tissue and low cartilage and muscle signal. The T2-FSE images had low signal for all tissues except synovial fluid which was bright. Despite the dramatic differences in tissue contrasts between the three image datasets, SUSAN was capable of translating the varying MR contrasts with a good visual appearance as a result of the incorporated CycleGAN functionality. The qualitative observation was supported by the quantitative evaluation. Figure \ref{fig5} further demonstrated an example of synthetic PD-FSE and T2-FSE images from a T1-SPGR image at different epochs with different FCN-scores calculated using ACC metric. In addition, there were FCN-scores (mean$\pm$standard deviation) of 0.73$\pm$0.03 and 0.79$\pm$0.05 for all the synthetic PD-FSE and T2-FSE images, and 0.78$\pm$0.02 and 0.83$\pm$0.04 for all the real hold-out PD-FSE and T2-FSE images, respectively, indicating a high similarity between the real images and the synthetic images from SUSAN. The influence of different weights of the segmentation branch on the image translation and segmentation accuracy are shown in Table \ref{tb2}. Although a large weight (${\lambda _{seg}}$=10) on the segmentation branch was not significantly different than the default moderate weight (${\lambda _{seg}}$=5) for translating and segmenting the PD-FSE and T2-FSE images, it was evident that a substantially reduced weight (${\lambda _{seg}}$=0.5) for the segmentation branch performed noticeably worse for the image translation and segmentation on both the PD-FSE and T2-FSE images.

\begin{table}[t]
  \caption{Comparison of image translation and segmentation accuracy (average value$\pm$standard deviation) of DC for SUSAN at different weights of the segmentation branch for the two clinical MR image datasets.}
   \label{tb2}
  \centering
  \begin{tabular}{llllllll}
    \toprule
    &&& \multicolumn{4}{c}{Dice Coefficient} \\
    \cmidrule{4-7}
    Dataset   &  ${\lambda _{seg}}$ & FCN-Score & Femur  & Tibia  & Femoral Cartilage & Tibial Cartilage  \\
    \midrule
    \multirow{3}{*}{PD-FSE}
     & 0.5  & 0.71$\pm$0.02   & 0.93$\pm$0.01   & 0.93$\pm$0.01   & 0.63$\pm$0.02   & 0.62$\pm$0.05 \\
     & 5  & 0.74$\pm$0.03   & 0.97$\pm$0.01  & 0.95$\pm$0.00   & 0.66$\pm$0.03   & 0.65$\pm$0.06 \\
     & 10  & 0.73$\pm$0.03   & 0.97$\pm$0.01   & 0.93$\pm$0.01   & 0.65$\pm$0.03   & 0.64$\pm$0.05 \\
    \cmidrule{2-7}
    \multirow{3}{*}{T2-FSE}    
    & 0.5  & 0.75$\pm$0.03  & 0.92$\pm$0.01   & 0.90$\pm$0.02   & 0.79$\pm$0.02   & 0.71$\pm$0.05 \\
    & 5  & 0.80$\pm$0.05   & 0.95$\pm$0.01   & 0.93$\pm$0.02   & 0.81$\pm$0.02   & 0.75$\pm$0.06 \\
    & 10  & 0.78$\pm$0.05   & 0.94$\pm$0.00   & 0.93$\pm$0.03   & 0.82$\pm$0.02   & 0.75$\pm$0.04 \\
    \bottomrule
  \end{tabular}
\end{table}

The averaged values (mean$\pm$standard deviation) of DC, VOE and ASSD are shown in Table \ref{tb3} for bone and cartilage segmentation for the 20 hold-out test subjects in the PD-FSE and T2-FSE image datasets. Although the segmentation accuracy was significantly higher for the supervised U-Net method than SUSAN for femoral cartilage (DC: \textit{p}=0.008, VOE: \textit{p}=0.008) on PD-FSE images and for tibia bone (DC: \textit{p}=0.008, ASSD: \textit{p}=0.002) on T2-FSE images. SUSAN provided overall comparable segmentation performance to the supervised U-Net method for the PD-FSE and T2-FSE image datasets while requiring no sequence specific annotated training data. The multi-atlas registration method performed significantly worse (\textit{p}<0.001 for all bone and cartilage) than both SUSAN and the supervised U-Net method. The direct registration method also performed significantly worse (\textit{p}<0.0001 for all bone and cartilage) than deep learning methods.

\begin{figure}[h]
  \centering
  \includegraphics[width=0.7\linewidth]{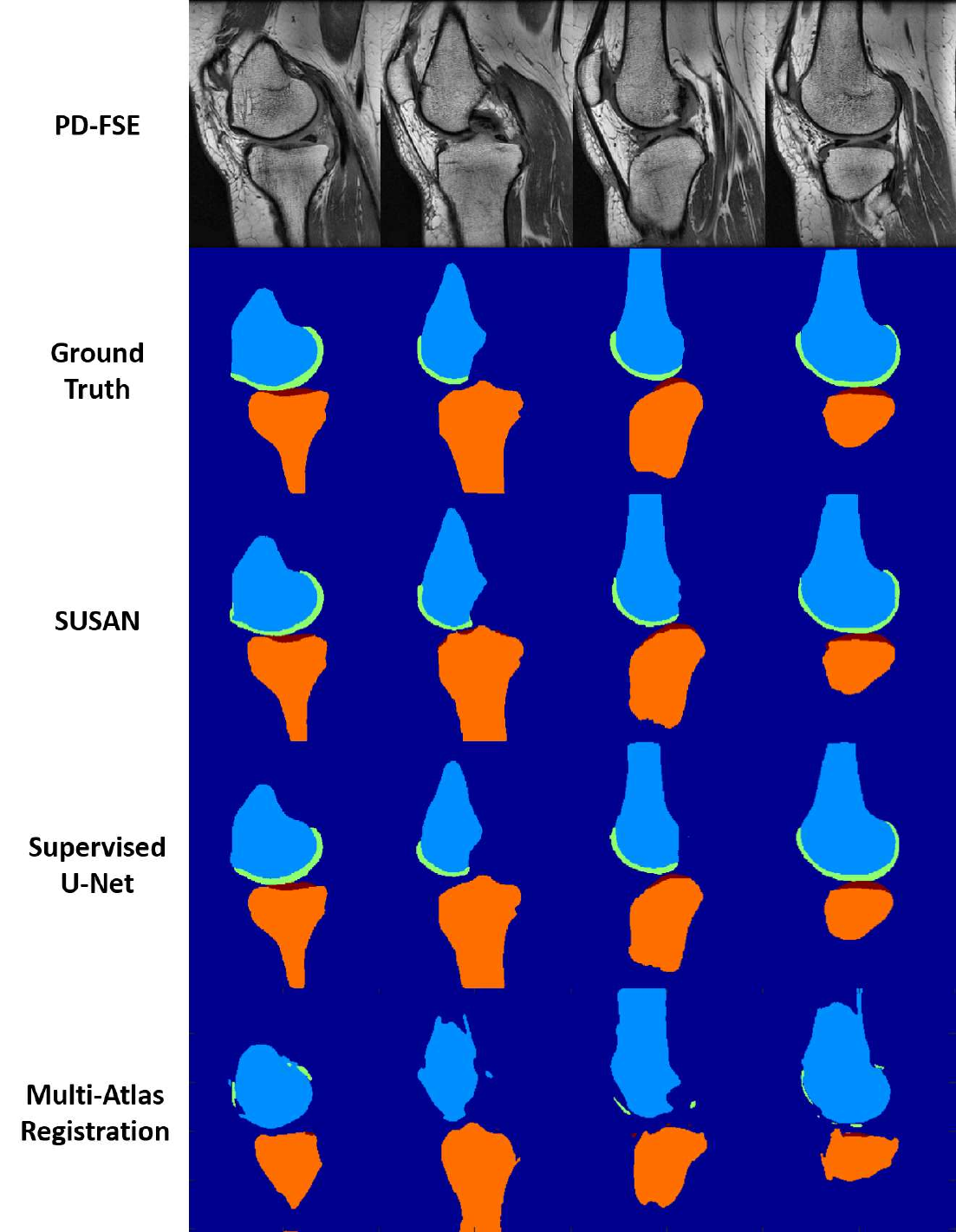}
  \caption{Example of bone and cartilage segmentation for a 56 year old male subject with mild knee osteoarthritis performed on the PD-FSE image dataset. SUSAN provided accurate segmentation relative to the ground truth with segmentation performance comparable to the supervised U-Net method. Both deep learning methods outperformed the multi-atlas registration method.}
 \label{fig6}
\end{figure}

Figure \ref{fig6} shows examples of bone and cartilage segmentation performed on sagittal PD-FSE images of the knee in a 56 year old male subject with mild knee osteoarthritis The segmentation results from SUSAN demonstrated good agreement with the overall contours of the ground truth. There was also good agreement of the overall shape between SUSAN and supervised U-Net method. For SUSAN, there was a bone segmentation accuracy of DC 0.94 and 0.95 for the femur and tibia respectively indicating small deviations from the ground truth. There was a cartilage segmentation accuracy of ASSD 0.84mm and 0.70mm for femoral and tibial cartilage respectively indicating good cartilage segmentation. For the supervised U-Net method, there was a bone segmentation accuracy of DC 0.94 and 0.96 for the femur and tibia respectively; there was a cartilage segmentation accuracy of ASSD 0.75mm and 0.65mm for femoral and tibial cartilage respectively. The deep learning methods outperformed the multi-atlas registration method. Although the multi-atlas method provided a similar bone shape to the ground truth for femur and tibia, the cartilage segmentation using the atlases built from SKI10 images failed on the PD-FSE images. For the direct registration, the segmentation result for both bone and cartilage was even worse than the multi-atlas registration and thus not shown in this figure.

\begin{figure}[h]
  \centering
  \includegraphics[width=0.7\linewidth]{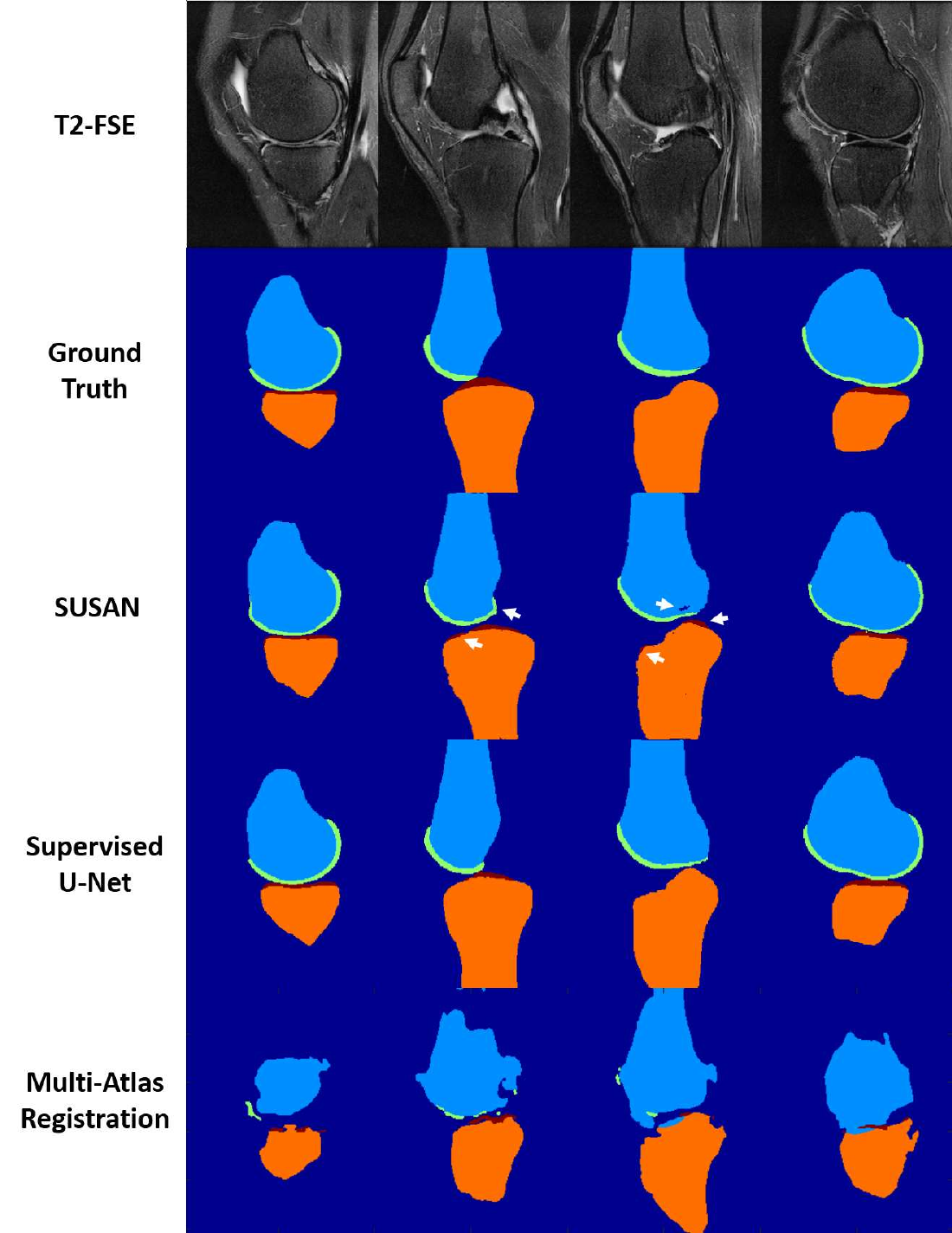}
  \caption{Example of bone and cartilage segmentation for a 64 year old male subject with severe knee osteoarthritis performed on the T2-FSE image dataset. Despite a few clusters of misclassification for bone and cartilage (white arrows), SUSAN provided good overall contours relative to the ground truth with segmentation performance comparable to the supervised U-Net method. Both deep learning methods outperformed the multi-atlas registration method.}
 \label{fig7}
\end{figure}

Figure \ref{fig7} shows examples of segmentation performed on sagittal T2-FSE images of the knee in a 64 year old male subject with knee osteoarthritis. Tissue segmentation was challenging for this subject due to the thin articular cartilage on the femur and tibia. The segmentation results from SUSAN demonstrated good agreement with the overall contours of the ground truth although a few small clusters of misclassification in bone and cartilage (white arrows) were observed due to ambiguous tissue contrast and low signal-to-noise ratio. For SUSAN, there was a bone segmentation accuracy of DC 0.96 and 0.93 for the femur and tibia respectively indicating small differences from the ground truth. There was a cartilage segmentation accuracy of ASSD 0.59mm and 0.71mm for femoral and tibial cartilage respectively indicating good cartilage segmentation. For the supervised U-Net method, there was a bone segmentation accuracy of DC 0.96 and 0.96 for the femur and tibia respectively; there was a cartilage segmentation accuracy of ASSD 0.46mm and 0.54mm for femoral and tibial cartilage respectively. The deep learning methods also substantially outperformed the multi-atlas registration method for bone and cartilage segmentation on T2-FSE images. The direct registration method failed to provide reasonable segmentation for both bone and cartilage for this subject thus not shown in this figure.

\begin{sidewaystable}[t]
  \setlength{\tabcolsep}{3pt}
  \caption{Comparison of segmentation accuracy (average value$\pm$standard deviation) of DC, VOE and ASSD for SUSAN, the supervised U-Net method, and two registration methods for the two clinical MR image datasets. The asterisks were used to indicate the significant differences (\textit{p}<0.05) of segmentation accuracy between the supervised U-Net method and SUSAN.  The multi-atlas registration method performed significantly worse (\textit{p}<0.001) than both SUSAN and the supervised U-Net method for all measures. The direct registration method also performed significantly worse (\textit{p}<0.0001) than both SUSAN and the supervised U-Net method for all measures.}
   \label{tb3}
  \centering
  \begin{tabular}{llllllllllllllll}
    \toprule
    && \multicolumn{2}{c}{Femur}  &&\multicolumn{2}{c}{Tibia} &&\multicolumn{3}{c}{Femoral Cartilage} &&\multicolumn{3}{c}{Tibial Cartilage}\\
    \cmidrule{3-4} \cmidrule{6-7} \cmidrule{9-11} \cmidrule{13-15} 
    Dataset   &  Method & DC & ASSD [mm]  && DC  & ASSD [mm] && DC & VOE & ASSD [mm] && DC & VOE & ASSD [mm]  \\
    \midrule
    \multirow{4}{*}{PD-FSE}
     & U-Net  & 0.96$\pm$0.01   & 1.29$\pm$0.31   && 0.96$\pm$0.01   & 0.83$\pm$0.43   && *0.77$\pm$0.01 &  *0.32$\pm$0.03  & 0.50$\pm$0.07    && 0.70$\pm$0.04 &  0.36$\pm$0.06  & 0.92$\pm$0.21 \\
     & SUSAN  & 0.97$\pm$0.01   & 1.32$\pm$0.25  && 0.95$\pm$0.00   & 0.90$\pm$0.16   && *0.66$\pm$0.03 &  *0.38$\pm$0.05 & 0.57$\pm$0.08 && 0.65$\pm$0.06 &  0.38$\pm$0.05 & 1.23$\pm$0.32\\
     & Multi-Atlas  & 0.86$\pm$0.10   & 5.11$\pm$2.11   && 0.80$\pm$0.15   & 4.63$\pm$3.15   && 0.25$\pm$0.13 &  0.85$\pm$0.11 & 5.16$\pm$3.07  && 0.36$\pm$0.21 &  0.82$\pm$0.15 & 6.32$\pm$2.46 \\
     & Direct-Registration  & 0.71$\pm$0.31   & 9.53$\pm$3.0   && 0.65$\pm$0.20   & 10.73$\pm$5.32  && 0.15$\pm$0.24 &  0.90$\pm$0.10 &  10.2$\pm$3.21  && 0.28$\pm$0.15 &  0.85$\pm$0.15 &  8.46$\pm$6.22 \\
    \cmidrule{3-4} \cmidrule{6-7} \cmidrule{9-11} \cmidrule{13-15} 
    \multirow{4}{*}{T2-FSE}    
     & U-Net  & 0.96$\pm$0.01   & 0.55$\pm$0.27   && *0.96$\pm$0.02   & *0.48$\pm$0.24   && 0.82$\pm$0.02 &  0.34$\pm$0.03  & 0.59$\pm$0.31    && 0.76$\pm$0.03 &  0.31$\pm$0.06  & 0.69$\pm$0.43 \\
     
     & SUSAN  & 0.95$\pm$0.01   & 0.65$\pm$0.20  && *0.93$\pm$0.02   & *1.1$\pm$0.28   && 0.81$\pm$0.02 &  0.35$\pm$0.04 & 0.65$\pm$0.31 && 0.75$\pm$0.06 &  0.33$\pm$0.06 & 0.73$\pm$0.36\\
     
     & Multi-Atlas  & 0.75$\pm$0.21   & 6.18$\pm$3.31   && 0.70$\pm$0.17   & 5.39$\pm$4.09   && 0.31$\pm$0.24 &  0.87$\pm$0.17 & 7.18$\pm$5.22  && 0.27$\pm$0.16 &  0.89$\pm$0.09 & 9.24$\pm$4.93 \\
     
     & Direct-Registration  & 0.65$\pm$0.18   & 8.92$\pm$4.46   && 0.53$\pm$0.21   & 8.15$\pm$4.39  && 0.26$\pm$0.15 &  0.88$\pm$0.18 &  8.29$\pm$3.13  && 0.25$\pm$0.14 &  0.89$\pm$0.13 &  10.16$\pm$4.52 \\

    \bottomrule
  \end{tabular}
\end{sidewaystable}

\section{DISCUSSION}
Our study described a novel adversarial CNN-based segmentation method that provided rapid and accurate segmentation for multiple MR image datasets using only a single set of annotated training data. This technique eliminates the need to retrain the segmentation CNNs using new annotation data specific to each MR sequence. The approach integrates the basic functions of the CycleGAN technique for image-to-image translation and utilizes an additional semantic segmentation network for joint image translation and segmentation. The segmentation results for segmenting two clinical knee MR image datasets suggested that this proposed method SUSAN utilizing joint adversarial and segmentation network can achieve high accuracy with performance comparable to a state-of-the-art supervised CNN method and better than registration methods. SUSAN was also highly time efficient with an average segmentation time less than one minute which is much shorter than the segmentation times of currently used fully-automated atlas-based and model-based segmentation methods.

A limitation of CNN-based segmentation methods is the need for adequate training data which may often be difficult to acquire. Many recent deep learning studies have demonstrated continuously  increasing segmentation performance with increasing training dataset size(58,59).  Since large training datasets are difficult to collect, several methods have been proposed to alleviate the problem. One approach is to implement data augmentation techniques where the size of the training data could be efficiently increased by applying random geometrical transformation to the original image data such as rotation, translation and deformation. These data augmentation techniques have become  useful techniques in many recent deep learning medical image studies to overcome small dataset size problem and to reduce the training overfitting(8,60). Transfer learning and fine-tuning techniques are other popular solutions to small training dataset size problem. Instead of initializing the network with random weights during training, the pre-trained weight values from the same image domain, different medical image domains, or even natural image datasets can be used to initiate the training process. Then, the training can be conducted in a supervised manner for the entire network or a few network layers using new annotated training data specific to the segmentation task. Many recent studies have shown that fine-tuning can improve performance and reduce requirement for training data and this performance improvement increases with reducing training dataset size(61,62). However, in both data augmentation and transfer learning techniques, annotated training data for the specific image dataset is still required, regardless of its size. We proposed a new approach to alleviate the problem by utilizing image translation to create domain specific training data. Our results suggest that a joint adversarial and segmentation network can be used for rapid and accurate segmentation for multiple MR image datasets using only a single set of annotated training data. Our method provides an alternative approach to perform medical image segmentation in circumstances in which collecting annotated training data is challenging, expensive, or not possible.

Although SUSAN was only evaluated for image translation for knee MR image datasets, there is no inherent limitation preventing this approach from translating images for other anatomic structures and for other imaging modalities. For example, it would be useful to translate knee joint cartilage segmentation into cartilage segmentation of the hip joint since annotating hip joint cartilage is extremely difficult due to the thin cartilage and closely opposing articular surfaces. One concern for such image translation is the significant geometric difference between image domains. Although the applied U-Net is widely accepted for mapping image contrast and texture as a results of efficient convolutional encoding and decoding design (46,47), successfully translation among different anatomy might still require further design of the generator architecture tailored for handling both contrast translation and geometric transformation. In addition, translating image contrasts among different imaging modalities would be useful in applications such as positron emission tomography (PET)/MR attenuation correction, where there is the need to generate synthetic computed tomography (CT) images from MR images for photon attenuation calculation (63,64). In current study, the use of adversarial training and cycle consistency regularization as key techniques in the CycleGAN framework was proven to be effective to learn mutually correlated image features with unpaired data in medical image domain. The adversarial learning ensures the translated images falling into the same data distribution of the target images; the cycle consistency prevents the degeneracy of the adversarial process from generating hallucinated image features (34).  This cross domain adaptation in CycleGAN imposes both shared-latent space constraints and information consistency to encourage CNNs to learn mutually correlated image features in different data distribution. Although current method achieves compelling results for translating MR images between different contrasts, the interpretation of the translated image contrast requires careful attentions in clinical practice. It should be noted that a typical assumption about high similarity of data characteristics in high dimensional feature space between training datasets might fail when pathology occurs in one contrast but not the other. Therefore, the synthetic images cannot be used as reliable diagnostic replacements for real images. Although the synthetic images in SUSAN provide sufficient information for image segmentation, they may not reflect true tissue contrasts for pathological conditions. Comprehensive assessment for the effectiveness and applicability of MR image translation using large prospective image datasets is needed for various diseases and tissue structures. In addition, SUSAN uses entirely unpaired image data thus might suffer from model collapse in adversarial training (34). Future comparison between current unpaired translation and the paired image-to-image translation which has stronger constraints in supervised learning is of great interests. Given a fraction of annotated data at minimal cost, a certain form of weakly supervised image-to-image translation incorporating both paired and unpaired information might provide further performance improvement.

Current SUSAN used 2D CNNs which can potentially limit image features within a single slice and may cause segmentation bias when contrast is inconsistent across slices in an image volume. This was the likely causes of the tissue misclassification noted in our results for the clinical knee MR images in Figure \ref{fig7} where tissue contrast was contaminated by the large noise level. Incorporating 3D information could be helpful for image translation to a target image with ambiguous tissue contrast. Multi-planner CNN methods, usually referred to as 2.5D methods, and fully 3D CNN methods have been shown in many studies to improve segmentation accuracy and efficiency (9,23,24). However, high dimensional CNN methods typically require extensive computing resources, such as abundant GPU memory, which is a prohibitively limiting factor in many studies. In our network configuration, the requirement for hardware resources when running high dimensional CNNs was even more severe since we implemented both forward and backward CNNs, which doubled the total network size. Alternatively, implementing suitable post-processing methods to adjust the segmentation results from a 2D CNN output is also applicable to take into account the 3D contextual relationships of the full image volume. Recent studies have demonstrated the successful use of fully-connected 3D conditional random field (CRF) to regularize segmentation boundaries at tissue interfaces (9,11,18). The 3D surface shape-based morphological deformable approach has also proven to be highly efficient to maintain desirable geometrical shape for segmented objects in combination with CNN-based segmentation method (17,18). Since these post-processing steps typical require no GPU computation, they could be very efficient and require little computational costs.

Our study has several limitations. First, the current study adapted the U-Net and PatchGAN structures and did not compare other CNN structures. Newly developed CNNs (65,66) will be explored in future studies. Second, the network training parameters were selected based on heuristic information. Tuning the weighting factors in the objective function is important to control the balance between image translation and structure segmentation, and is likely dependent of specific studies. Although the results from this study demonstrated that a moderately weighted segmentation branch improved the image translation (Table \ref{tb2}), comprehensive parameter optimization would be necessary in future studies to investigate further improvement for the performance. Third, future studies investigating the influence of imaging parameters on the image translation and segmentation accuracy are also needed. Finally, the current method did not compare other methods which use adversarial networks for image segmentation.

\section{CONCLUSIONS}
In conclusion, our study described a new fully-automated CNN-based segmentation method which integrates joint adversarial and segmentation CNNs to segment MR images with different tissue contrasts using a single set of annotated training data. Our method was shown to provide rapid and accurate segmentation of bone and cartilage for clinical knee MR image datasets comparable to a state-of-the-art supervised CNN method. Additional studies are needed to evaluate potential applications of SUSAN for other anatomical structures and for other imaging modalities. The new technique may further improve the applicability and efficiency of CNN-based segmentation of medical images while eliminating the need for large amounts of annotated training data.

\section*{REFERENCES}
\medskip
\small
1. Yushkevich PA, Piven J, Hazlett HC, Smith RG, Ho S, Gee JC, Gerig G. User-guided 3D active contour segmentation of anatomical structures: significantly improved efficiency and reliability. Neuroimage 2006;31:1116–1128.

2. McWalter EJ, Wirth W, Siebert M, von Eisenhart-Rothe RMO, Hudelmaier M, Wilson DR, Eckstein F. Use of novel interactive input devices for segmentation of articular cartilage from magnetic resonance images. Osteoarthr. Cartil. [Internet] 2005;13:48–53. doi: 10.1016/j.joca.2004.09.008.

3. Ashton EA, Takahashi C, Berg MJ, Goodman A, Totterman S, Ekholm S. Accuracy and reproducibility of manual and semiautomated quantification of MS lesions by MRI. J. Magn. Reson. Imaging [Internet] 2003;17:300–308. doi: 10.1002/jmri.10258.

4. Shim H, Chang S, Tao C, Wang J-H, Kwoh CK, Bae KT. Knee Cartilage: Efficient and Reproducible Segmentation on High-Spatial-Resolution MR Images with the Semiautomated Graph-Cut Algorithm Method. Radiology [Internet] 2009;251:548–556. doi: 10.1148/radiol.2512081332.

5. Sharma N, Aggarwal LM. Automated medical image segmentation techniques. J. Med. Phys. [Internet] 2010;35:3–14. doi: 10.4103/0971-6203.58777.

6. Pedoia V, Majumdar S, Link TM. Segmentation of joint and musculoskeletal tissue in the study of arthritis. Magn. Reson. Mater. Physics, Biol. Med. [Internet] 2016;29:207–221. doi: 10.1007/s10334-016-0532-9.

7. Litjens G, Kooi T, Bejnordi BE, Setio AAA, Ciompi F, Ghafoorian M, van der Laak JAWM, van Ginneken B, Sánchez CI. A survey on deep learning in medical image analysis. Med. Image Anal. 2017;42:60–88. doi: 10.1016/j.media.2017.07.005.

8. Pereira S, Pinto A, Alves V, Silva CA. Brain Tumor Segmentation using Convolutional Neural Networks in MRI Images. IEEE Trans Med Imaging [Internet] 2016. doi: 10.1109/TMI.2016.2538465.

9. Kamnitsas K, Ledig C, Newcombe VFJ, Simpson JP, Kane AD, Menon DK, Rueckert D, Glocker B. Efficient multi-scale 3D CNN with fully connected CRF for accurate brain lesion segmentation. Med. Image Anal. 2017;36:61–78. doi: 10.1016/j.media.2016.10.004.

10. Moeskops P, Viergever MA, Mendrik AM, de Vries LS, Benders MJ, Isgum I. Automatic segmentation of MR brain images with a convolutional neural network. IEEE Trans Med Imaging [Internet] 2016. doi: 10.1109/TMI.2016.2548501.

11. Zhao G, Liu F, Oler JA, Meyerand ME, Kalin NH, Birn RM. Bayesian convolutional neural network based MRI brain extraction on nonhuman primates. Neuroimage [Internet] 2018;175:32–44. doi: 10.1016/j.neuroimage.2018.03.065.

12. Brosch T, Tang L, Yoo Y, Li D, Traboulsee A, Tam R. Deep 3D Convolutional Encoder Networks with Shortcuts for Multiscale Feature Integration Applied to Multiple Sclerosis Lesion Segmentation. IEEE Trans Med Imaging [Internet] 2016. doi: 10.1109/TMI.2016.2528821.

13. Avendi MR, Kheradvar A, Jafarkhani H. Fully automatic segmentation of heart chambers in cardiac MRI using deep learning. J. Cardiovasc. Magn. Reson. [Internet] 2016;18:2–4. doi: 10.1186/1532-429X-18-S1-P351.

14. Avendi MR, Kheradvar A, Jafarkhani H. Automatic segmentation of the right ventricle from cardiac MRI using a learning-based approach. Magn. Reson. Med. [Internet] 2017;78:2439–2448. doi: 10.1002/mrm.26631.

15. Ferdinand Christ P, Ezzeldin Elshaer MA, Ettlinger F, et al. Automatic Liver and Lesion Segmentation in CT Using Cascaded Fully Convolutional Neural Networks and 3D Conditional Random Fields. Proc. 19th Int. Conf. Med. Image Comput. Comput. Assist. Interv. [Internet] 2016:1–8. doi: 10.1007/978-3-319-46723-8.

16. Zha W, Kruger SJ, Johnson KM, Cadman R V., Bell LC, Liu F, Hahn AD, Evans MD, Nagle SK, Fain SB. Pulmonary ventilation imaging in asthma and cystic fibrosis using oxygen-enhanced 3D radial ultrashort echo time MRI. J. Magn. Reson. Imaging [Internet] 2017. doi: 10.1002/jmri.25877.

17. Liu F, Zhou Z, Jang H, Samsonov A, Zhao G, Kijowski R. Deep Convolutional Neural Network and 3D Deformable Approach for Tissue Segmentation in Musculoskeletal Magnetic Resonance Imaging. Magn. Reson. Med. [Internet] 2017:DOI: 10.1002/mrm.26841. doi: 10.1002/mrm.26841.

18. Zhou Z, Zhao G, Kijowski R, Liu F. Deep Convolutional Neural Network for Segmentation of Knee Joint Anatomy. Magn. Reson. Med. 2018:doi:10.1002/mrm.27229.

19. Liu F, Zhou Z, Samsonov A, Blankenbaker D, Larison W, Kanarek A, Lian K, Kambhampati S, Kijowski R. Deep Learning Approach for Evaluating Knee MR Images: Achieving High Diagnostic Performance for Cartilage Lesion Detection. Radiology [Internet] 2018:172986. doi: 10.1148/radiol.2018172986.

20. Ronneberger O, Fischer P, Brox T. U-Net: Convolutional Networks for Biomedical Image Segmentation. In: Navab N, Hornegger J, Wells WM, Frangi AF, editors. Medical Image Computing and Computer-Assisted Intervention MICCAI 2015: 18th International Conference, Munich, Germany, October 5-9, 2015, Proceedings, Part III. Cham: Springer International Publishing; 2015. pp. 234-241. doi: 10.1007/978-3-319-24574-4\_28.

21. Badrinarayanan V, Kendall A, Cipolla R. SegNet: A Deep Convolutional Encoder-Decoder Architecture for Image Segmentation. ArXiv e-prints [Internet] 2015.

22. Guerra E, de Lara J, Malizia A, Díaz P. Supporting user-oriented analysis for multi-view domain-specific visual languages. Inf. Softw. Technol. [Internet] 2009;51:769–784. doi: 10.1016/j.infsof.2008.09.005.

23. Milletari F, Navab N, Ahmadi S-A. V-Net: Fully Convolutional Neural Networks for Volumetric Medical Image Segmentation. ArXiv e-prints [Internet] 2016.

24. Çiçek Ö, Abdulkadir A, Lienkamp SS, Brox T, Ronneberger O. 3D U-Net: Learning Dense Volumetric Segmentation from Sparse Annotation. ArXiv e-prints [Internet] 2016.

25. Wachinger C, Reuter M, Klein T. DeepNAT: Deep Convolutional Neural Network for Segmenting Neuroanatomy. ArXiv e-prints [Internet] 2017. doi: 10.1016/j.neuroimage.2017.02.035.

26. Dolz J, Desrosiers C, Ben Ayed I. 3D fully convolutional networks for subcortical segmentation in MRI: A large-scale study. Neuroimage [Internet] 2017. doi: 10.1016/J.NEUROIMAGE.2017.04.039.

27. Luc P, Couprie C, Chintala S, Verbeek J, Kuntzmann LJ. Semantic Segmentation using Adversarial Networks. Arxiv [Internet] 2016.

28. Kamnitsas K, Baumgartner C, Ledig C, et al. Unsupervised domain adaptation in brain lesion segmentation with adversarial networks. ArXiv e-prints [Internet] 2016.

29. Moeskops P, Veta M, Lafarge MW, Eppenhof KAJ, Pluim JPW. Adversarial Training and Dilated Convolutions for Brain MRI Segmentation. In: Deep Learning in Medical Image Analysis and Multimodal Learning for Clinical Decision Support. Springer, Cham; 2017. pp. 56–64. doi: 10.1007/978-3-319-67558-9\_7.

30. Kohl S, Bonekamp D, Schlemmer H-P, Yaqubi K, Hohenfellner M, Hadaschik B, Radtke J-P, Maier-Hein K. Adversarial Networks for the Detection of Aggressive Prostate Cancer. arXiv Prepr. arXiv1702.08014 [Internet] 2017.

31. Dai W, Doyle J, Liang X, Zhang H, Dong N, Li Y, Xing EP. SCAN: Structure Correcting Adversarial Network for Organ Segmentation in Chest X-rays. ArXiv e-prints [Internet] 2017.

32. Lafarge MW, Pluim JPW, Eppenhof KAJ, Moeskops P, Veta M. Domain-Adversarial Neural Networks to Address the Appearance Variability of Histopathology Images. In: Springer, Cham; 2017. pp. 83–91. doi: 10.1007/978-3-319-67558-9\_10.

33. Greenspan  Bram van HG, Summers RM, Greenspan H, et al. Guest Editorial Deep Learning in Medical Imaging: Overview and Future Promise of an Exciting New Technique. IEEE Trans. Med. Imaging [Internet] 2016;35:1153–1159. doi: 10.1109/TMI.2016.2553401.

34. Zhu J-Y, Park T, Isola P, Efros AA. Unpaired Image-to-Image Translation Using Cycle-Consistent Adversarial Networks. In: 2017 IEEE International Conference on Computer Vision (ICCV). Vol. 2017–Octob. IEEE; 2017. pp. 2242–2251. doi: 10.1109/ICCV.2017.244.

35. Brislin RW. Back-Translation for Cross-Cultural Research. J. Cross. Cult. Psychol. [Internet] 1970;1:185–216. doi: 10.1177/135910457000100301.

36. Xia Y, He D, Qin T, Wang L, Yu N, Liu T-Y, Ma W-Y. Dual Learning for Machine Translation. ArXiv e-prints [Internet] 2016.

37. Zhou T, Krähenbühl P, Aubry M, Huang Q, Efros AA. Learning Dense Correspondence via 3D-guided Cycle Consistency. ArXiv e-prints [Internet] 2016.

38. Wang F, Huang Q, Guibas LJ. Image Co-segmentation via Consistent Functional Maps. In: 2013 IEEE International Conference on Computer Vision. IEEE; 2013. pp. 849–856. doi: 10.1109/ICCV.2013.110.

39. Huang Q-X, Guibas L. Consistent Shape Maps via Semidefinite Programming. Comput. Graph. Forum [Internet] 2013;32:177–186. doi: 10.1111/cgf.12184.

40. Pathak D, Krahenbuhl P, Donahue J, Darrell T, Efros AA. Context Encoders: Feature Learning by Inpainting. ArXiv e-prints [Internet] 2016.

41. Reed S, Akata Z, Yan X, Logeswaran L, Schiele B, Lee H. Generative Adversarial Text to Image Synthesis. ArXiv e-prints [Internet] 2016.

42. Denton E, Chintala S, Szlam A, Fergus R. Deep Generative Image Models using a Laplacian Pyramid of Adversarial Networks. ArXiv e-prints [Internet] 2015.

43. Radford A, Metz L, Chintala S. Unsupervised Representation Learning with Deep Convolutional Generative Adversarial Networks. ArXiv e-prints [Internet] 2015.

44. Goodfellow IJI, Pouget-Abadie J, Mirza M, Xu B, Warde-Farley D, Ozair S, Courville A, Bengio Y. Generative Adversarial Networks. arXiv Prepr. arXiv … [Internet] 2014:1–9. doi: 10.1001/jamainternmed.2016.8245.

45. Long J, Shelhamer E, Darrell T. Fully Convolutional Networks for Semantic Segmentation. ArXiv e-prints [Internet] 2014;1411.

46. Isola P, Zhu J-Y, Zhou T, Efros AA. Image-to-Image Translation with Conditional Adversarial Networks. ArXiv e-prints [Internet] 2016.

47. Gong E, Pauly JM, Wintermark M, Zaharchuk G. Deep learning enables reduced gadolinium dose for contrast-enhanced brain MRI. J. Magn. Reson. Imaging [Internet] 2018. doi: 10.1002/jmri.25970.

48. Jia Y, Shelhamer E, Donahue J, Karayev S, Long J, Girshick R, Guadarrama S, Darrell T. Caffe: Convolutional Architecture for Fast Feature Embedding. ArXiv e-prints [Internet] 2014;1408.

49. Ledig C, Theis L, Huszar F, et al. Photo-Realistic Single Image Super-Resolution Using a Generative Adversarial Network. ArXiv e-prints [Internet] 2016.

50. Li C, Wand M. Precomputed Real-Time Texture Synthesis with Markovian Generative Adversarial Networks. ArXiv e-prints [Internet] 2016.

51. Heimann T, Morrison B, Styner M, Niethammer M, Warfield S. Segmentation of Knee Images: A Grand Challenge. Med Image Comput Comput Assist Interv 2010:207–214. doi: citeulike-article-id:12455897.

52. He K, Zhang X, Ren S, Sun J. Delving Deep into Rectifiers: Surpassing Human-Level Performance on ImageNet Classification. ArXiv e-prints [Internet] 2015;1502.

53. Kingma DP, Ba J. Adam: A Method for Stochastic Optimization. ArXiv e-prints [Internet] 2014.

54. Shan L, Zach C, Charles C, Niethammer M. Automatic atlas-based three-label cartilage segmentation from MR knee images. Med. Image Anal. [Internet] 2014;18:1233–1246. doi: 10.1016/j.media.2014.05.008.

55. Bron EE, van Tiel J, Smit H, Poot DHJ, Niessen WJ, Krestin GP, Weinans H, Oei EHG, Kotek G, Klein S. Image registration improves human knee cartilage T1 mapping with delayed gadolinium-enhanced MRI of cartilage (dGEMRIC). Eur. Radiol. [Internet] 2013;23:246–52. doi: 10.1007/s00330-012-2590-3.

56. François Chollet. Keras. GitHub 2015:https://github.com/fchollet/keras.

57. Abadi M, Agarwal A, Barham P, et al. TensorFlow: Large-Scale Machine Learning on Heterogeneous Distributed Systems. ArXiv e-prints [Internet] 2016. doi: 10.1109/TIP.2003.819861.

58. Cho J, Lee K, Shin E, Choy G, Do S. How much data is needed to train a medical image deep learning system to achieve necessary high accuracy? ArXiv e-prints [Internet] 2015.

59. Lekadir K, Galimzianova A, Betriu A, del Mar Vila M, Igual L, Rubin DL, Fernandez E, Radeva P, Napel S. A Convolutional Neural Network for Automatic Characterization of Plaque Composition in Carotid Ultrasound. IEEE J. Biomed. Heal. Informatics [Internet] 2017;21:48–55. doi: 10.1109/JBHI.2016.2631401.

60. Krizhevsky A, Sutskever I, Hinton GE. ImageNet Classification with Deep Convolutional Neural Networks. Adv. Neural Inf. Process. Syst. 25 [Internet] 2012:1097–1105.

61. Shin H-C, Roth HR, Gao M, Lu L, Xu Z, Nogues I, Yao J, Mollura D, Summers RM. Deep Convolutional Neural Networks for Computer-Aided Detection: CNN Architectures, Dataset Characteristics and Transfer Learning. ArXiv e-prints [Internet] 2016;1602.

62. Tajbakhsh N, Shin JY, Gurudu SR, Hurst RT, Kendall CB, Gotway MB, Liang J. Convolutional Neural Networks for Medical Image Analysis: Full Training or Fine Tuning? IEEE Trans. Med. Imaging [Internet] 2016;35:1299–1312. doi: 10.1109/TMI.2016.2535302.

63. Liu F, Jang H, Kijowski R, Bradshaw T, McMillan AB. Deep Learning MR Imaging–based Attenuation Correction for PET/MR Imaging. Radiology [Internet] 2017:170700. doi: 10.1148/radiol.2017170700.

64. Jang H, Liu F, Zhao G, Bradshaw T, McMillan AB. Technical Note: Deep learning based MRAC using rapid ultra-short echo time imaging. Med. Phys. [Internet] 2018:In-press. doi: 10.1002/mp.12964.

65. Mirza M, Osindero S. Conditional Generative Adversarial Nets. ArXiv e-prints [Internet] 2014.

66. Karras T, Aila T, Laine S, Lehtinen J. Progressive Growing of GANs for Improved Quality, Stability, and Variation. ArXiv e-prints [Internet] 2017.
\end{document}